\RequirePackage[normalem]{ulem} 
\RequirePackage{color}\definecolor{RED}{rgb}{1,0,0}\definecolor{BLUE}{rgb}{0,0,1} 

 \documentclass[tablecaption=bottom,wcp]{jmlr} 

\usepackage{booktabs}
\usepackage{ulem}

\usepackage[load-configurations=version-1]{siunitx} 
\usepackage{enumerate}
\usepackage[capitalize]{cleveref}
\usepackage{adjustbox}

\theorembodyfont{\upshape}
\theoremheaderfont{\scshape}
\theorempostheader{:}
\theoremsep{\newline}

\DeclareMathOperator{\EX}{\mathbb{E}}

\jmlrproceedings{AABI 2023}{5th Symposium on Advances in Approximate Bayesian Inference, 2023}

\title[Uncertainty in Graph Contrastive Learning with BNNs]{Uncertainty in Graph Contrastive Learning \\ with Bayesian Neural Networks \vspace{0.6em}}

 \author{\Name{Alexander Möllers}\nametag{\thanks{Work done while at ETH Zürich. Correspondence to a.j.moellers@gmail.com}} \\
 \addr TU Delft, Delft, The Netherlands
 \AND
 \Name{Alexander Immer} \\
 \addr ETH Zürich, Zürich, Switzerland
 \AND
 \Name{Elvin Isufi} \\
 \addr TU Delft, Delft, The Netherlands
 \AND
 \Name{Vincent Fortuin} \\
 \addr Helmholtz AI, Munich, Germany
 }

\begin{document}

\maketitle

\begin{abstract}
    Graph contrastive learning has shown great promise when labeled data is scarce, but large unlabeled datasets are available. However, it often does not take uncertainty estimation into account. We show that a variational Bayesian  neural network approach can be used to improve not only the uncertainty estimates but also the downstream performance on semi-supervised node-classification tasks. Moreover, we propose a new measure of uncertainty for contrastive learning, that is based on the disagreement in likelihood due to different positive samples.
\end{abstract}


\section{Introduction}
\label{sec:intro}

Traditional supervised learning with deep neural networks does not leverage the information contained in unlabeled data and performs poorly in scarce data settings. Partially as a response to this, self-supervised training methods for neural networks on graphs and images have seen great progress in recent years. Hereby, the InfoNCE objective introduced by \citet{Oord2018} has played an important part as it constitutes the basis for many of the best-performing methods that have been developed \citep{SimCLR,MOCO,GraphCL,Grace}. This contrastive loss uses positive and negative examples for each data point and encourages the network to learn a function that maps the positives close to each other and the negatives further apart in an embedding space. Often, the positive samples are generated by creating augmentations of the original data point and the negative ones are taken to be random samples from the data.  \\

For many application areas of contrastive learning, such as object detection \citep{obdet} or drug discovery \citep{sanchez-fernandez2022contrastive}, obtaining accurate uncertainty estimates is essential. Nevertheless, research into probabilistic approaches to InfoNCE learning is limited. A notable exception is \citet{aitchison2023infonce}, who have shown that the weights of the encoding neural network (encoder) can be learned by optimizing an evidence lower bound (ELBO). We extend this result to include distributions over the parameters and equip the model with a notion of (Bayesian) epistemic uncertainty. Related to this, we make the following contributions: 

\begin{enumerate}[(i)]
  \item We propose variational graph constrastive learning (VGCL) and find that regularizing the variational family leads to a remarkable improvement in downstream accuracy. 
  \item We investigate the uncertainty calibration on a downstream task with respect to different existing uncertainty measures. We find all the measures to be improved when VGCL is used and weight uncertainty is taken into account.   
  \item We propose the contrastive model disagreement score (CMDS), a new approach for measuring uncertainty in contrastive learning based on the disagreement between positive samples.  We empirically show that it outperforms the currently used measures. 
\end{enumerate}

\section{Background}

\subsection{Deep Learning on Graphs}

Deep learning on graphs aims to develop neural network models for graph-structured data. Graph Neural Networks (GNNs) leverage convolutional filters to process the information contained in nodes and edges to extract meaningful features and predict graph properties \citep{gama2020graphs,wu2020survey}. The problem setting which we investigate is node classification. Here, the observed data is one input graph consisting of $n$ nodes with node features $\vec{X} \in \mathbb{R}^{n \times f}$ and an adjacency matrix $\vec{A} \in \mathbb{R}^{n \times n}$. In the self-supervised setting, we aim to train a network that maps the observed data on this manifold into the Euclidean space $\mathbb{R}^{m}$ in a way that results in a useful node embedding. The quality of these embeddings is evaluated via a linear classifier trained on the node labels.

\subsection{Self-Supervised Learning with the InfoNCE}

The InfoNCE \citep{Oord2018} considers two data items $x$ and $x'$, where $x$ could be the original data point or anchor and $x'$ could be an augmentation of $x$. These data points are mapped by an encoder $\mathrm{Q}_\phi(z \mid x)$ to their respective embeddings $z$ and $z'$. Here, $\phi$  are the parameters of the encoder. The InfoNCE objective is then defined as:

\begin{equation} \label{eqn:infonce}
 \mathrm{I}_{N}(\phi,	\rho ) = \EX_{X}\left(\log\frac{f_{\rho }(z,z')}{f_{\rho}(z,z')+ \sum_{j=1}^{N}f_{\rho}(z,z_{j})}\right) +\log(N)
\end{equation}

where $z_{j}$ are the embeddings of $N$ negative examples and $f_{\rho}(\cdot,\cdot)$ is a similarity function, parameterised by $\rho$, that is used to distinguish the samples in the embedding space. Usually it is taken to be the cosine similarity coupled with an additional MLP layer called the projection head. 

\subsection{A Probabilistic Interpretation of the InfoNCE}

Recently, \citet{aitchison2023infonce} have shown that contrastive learning can be interpreted as a generative process and have formulated a related statistical model. In this setting, pairs of data points $x$, $x'$ are observed that are generated by the latent variables $z$ and $z'$. The likelihood $\mathrm{P}_{\phi,\rho}\left(x, x^{\prime}\right)$ is a joint distribution over the observed data and it is parametrised by ($\phi$, $\rho$).  The model parameters that maximize the corresponding model evidence can then be approximately found by optimizing the following evidence lower bound (further details in \cref{apd:der elbo}): 

\begin{equation}
\log \mathrm{P}_{\phi,\rho}\left(x, x^{\prime}\right)\geq \mathbb{E}_{\mathrm{Q}_{\phi}\left(z, z^{\prime}\mid x,x^{\prime}\right)}\left[\log\frac{\mathrm{P}_{\theta,\rho}\left(z,z^{\prime}\right)}{\mathrm{Q}_\phi\left(z^{\prime}\right)\mathrm{Q}_{\phi}\left(z\right)}\right] + \log \mathrm{P}_{true}(x)\mathrm{P}_{true}(x')
\end{equation}

where $\mathrm{Q}_\phi\left(z\right)$ and $\mathrm{Q}_\phi\left(z^{\prime}\right)$ are the marginals of the joint distribution $\mathrm{Q}_{\phi}\left(z, z^{\prime}\right)$. Hereby, the true distributions do not depend on the parameters of the model and we can optimize the bound without taking them into consideration. Moreover, $\mathrm{P}_{\phi,\rho}\left(z,z^{\prime}\right)$ is a prior over the latent variables. It turns out that when we choose it as $\mathrm{P}_{\phi,\rho}\left(z,z^{\prime}\right) = \mathrm{P}_{\phi,\rho}\left(z\right) \, \mathrm{P}_{\phi,\rho}\left(z^{\prime} \mid z \right)$ and pick:

\begin{equation}
\label{eqn:4}
\begin{aligned}
\mathrm{P}_{\phi,\rho}(z) & =\mathrm{Q}_\phi(z) \\
\mathrm{P}_{\phi, \rho}\left(z^{\prime} \mid z\right) & =\frac{1}{Z_{\phi, \rho}(z)} \mathrm{Q}_\phi\left(z^{\prime}\right) f_\rho\left(z, z^{\prime}\right)
\end{aligned}
\end{equation}

with $Z_{ \phi,\rho}(z)$ being a normalizing constant, then we obtain an objective that is equivalent to the InfoNCE (up to a constant) for a deterministic encoder (\cref{apd:InfoNCE_ELBO}). A more detailed treatment of the related work is deferred to \cref{sec:related_work}.

\section{Methods}

\subsection{VGCL: Contrastive Learning with Probabilistic Encoders}

To obtain a model that accounts for the epistemic uncertainty in the parameters, we alter the formulation of the evidence lower bound for contrastive learning to include a distribution over the weights $w = (\phi,\rho)$. This yields a Variational Graph Contrastive Learning (VGCL) approach. The related ELBO (\cref{apd:der elbo}) that we would like to optimize is:

\begin{align}
\log \mathrm{P}\left(x, x^{\prime}\right)\geq \mathbb{E}_{\mathrm{Q} \left(z, z^{\prime} \mid x,x^{\prime},w \right)\mathrm{q} (w\mid \theta)}\left[ \log \frac{\mathrm{P} \left(z,z^{\prime} \mid w \right)}{\mathrm{Q}\left(z^{\prime}\mid w\right)\mathrm{Q}\left(z' \mid w \right)} \right] - KL(\mathrm{q}(w\mid \theta) \, \| \, \mathrm{P}(w))
\end{align}

where the last term is the Kullback–Leibler divergence between the prior over the weights $P(w)$ and the variational family $q(w\mid \mathbf{\theta})$ parametrised by $\theta$. For our experiments, we pick them to be Gaussian distributions. We thus have $P(w)=\prod_j \mathcal{N}\left(w_j \mid \mu , \sigma^2\right)$ and $q(w\mid \mathbf{\theta} )=\prod_j \mathcal{N}\left(w_j \mid \mathbf{\theta}\right)$ where each $w_{j}$ is a weight of the neural network. Furthermore, following \citet{blundell2015bbb}, we reparameterize the standard deviation of the variational distribution as $\sigma_{v}=\log (1+\exp (p))$ and the parameter set is thus $\theta=(\mu,p)$. In addition to that, we regularize the variational family by placing Gaussian hyperpriors over the variational parameters:

\begin{equation}
 \mu_{v} \sim \mathcal{N}(0, \sigma_{0}^{2}) \, , \, p \sim \mathcal{N}(\mu_{p}, \sigma_{p}^{2})
\end{equation}

This is meant to encourage a larger variance in the weights, due to the fact that we are training the model on augmentations, which are only an approximation to the data that would be the result of a generating process given the true latents. We thus hypothesize that a larger uncertainty about the model weights might be beneficial and we incorporate this prior knowledge by regularizing the variational family. 

\subsection{CMDS: Uncertainty in Contrastive Learning}
\label{sec:cmds}

Existing works on uncertainty in contrastive learning either do not model the uncertainty related to a data point $x$ explicitly or focus on its embedding uncertainty (\cref{sec:related_work}). That is, they use a trained encoder to map $x$ into the embedding space and then reason about the variance of the embedding features (e.g. under different augmentations). Instead, we use the probabilistic model of the InfoNCE to design a measure of uncertainty that is related to the likelihood of the data and incorporates the epistemic uncertainty of the weight distribution when using a BNN. To this end, we note that the likelihood in the probabilistic model of contrastive learning is the contrastive loss and that it is calculated based on $x$ and $x'$. Thus, in contrast to other works, we encode two data points and reason about the quantity $\mathrm{P}_{\theta, \rho}(x,x')$. 

\begin{table}
\centering
  \caption{Test accuracies for different unsupervised methods on the citation datasets. Our proposed method outperforms all the baselines on all tasks.}
  \label{tab:acc_table}
  \scalebox{0.85}{
  \begin{tabular}{llll}
  \toprule
  \textbf{Dataset} &  \textbf{Cora} &  \textbf{Citeseer} &  \textbf{Pubmed}\\
  \midrule
  Raw features & $64.8$ & $64.6$ & $84.8$\\
  node2vec & 74.8 & 52.3 & $80.3$\\
  DeepWalk & 75.7 & 50.5 & $80.5$\\
  Deep Walk + Features & $73.1$ & $47.6$ & $83.7$\\
  \midrule
  GAE & $76.9$ & $60.6$ & $82.9$\\
  VGAE & $78.9$ & $61.2$ & $83.0$\\
  DGI  & $82.6 \pm 0.4$ & $68.8 \pm 0.7$ & $86.0 \pm 0.1$\\
  InfoNCE & $81.9 \pm 0.4 $ & $70.8 \pm 0.1$ & $85.0 \pm 0.1$\\
  VI-InfoNCE & $82.1 \pm 0.3 $ & $71.0 \pm 0.1$ & $85.0 \pm 0.1$\\
\midrule
 VGCL (Ours) & $\mathbf{83.5 \pm 0.2}$ & $\mathbf{72.2 \pm 0.1}$ & $\mathbf{86.3 \pm 0.1}$\\
  \bottomrule
  \end{tabular}}
\end{table}

To incorporate the contrastive nature of the problem, we propose to quantify the uncertainty of $x$ by using the variation of $\mathrm{P}_{\theta, \rho}(x\mid x')$  under samples $x' \sim p(x' \mid z')$. The intuition is that for a data point and context for which the model has learned a coherent explanation, it will assign similar likelihoods for different positive examples. The samples can be obtained by creating augmentations of $x$.

Notably, a related idea exists in out-of-distribution detection for Bayesian Variational Autoencoders, where \citet{BVAE} measure the disagreement in the likelihood estimates of different models sampled from a learned weight distribution. Extending their idea to contrastive learning, we calculate the related disagreement score not only for weights but also for positive samples. This leads to the Contrastive Model Disagreement Score (CMDS): 
 \begin{align}
D_{\text{CMDS}}\left(x\right)=\frac{1}{M\sum_{j=1}^{M} l_{j}(x)^2} \,  \qquad \text{with} \quad l_{j}(x)=\frac{\mathrm{P}\left(x \mid x'_{j}, \phi_{j}, \rho_{j} \right)}{\sum_{j=1}^{M} \mathrm{P}\left(x \mid x'_{j}, \phi_{j}, \rho_{j} \right)},
\end{align}
where we generate $M$ samples from $x'_{j} \sim  p(x\mid z')$ via augmentations and $\rho_{j}, \phi_{j} \sim q(w\mid \theta)$ from the learned variational families. This score will be large when the normalised likelihoods $l_{j}$ are alike and small when they differ a lot. It is important to emphasize that, while the mathematical formulation of the CMDS is similar to the measure proposed by \citet{BVAE} there are some fundamental differences. Importantly, the CMDS can be applied to deterministic models and it is based on the joint likelihood between $x,x'$ that is inherent to contrastive learning. The measure proposed by \citet{BVAE} can only be used for Bayesian models and it is defined for the setting where a Variational Autoencoder is trained with a reconstruction loss. There, the likelihood is only based on one data point. For an extended discussion and mathematical explanation of the CMDS we defer the reader to \cref{apd:Dis Score,apd:Uncertainty Measure}.

\section{Experiments}

\begin{figure}[t]
\floatconts
  {fig:tsns}
  {\caption{T-SNE plots of the embeddings generated by a deterministic encoder (left) in comparison to the embeddings generated by our probabilistic encoder with hyperpriors (right). Our method generates a better separation between the classes in the embedding spaces, which might explain its improved performance.}}
  {\includegraphics[width=1\linewidth]{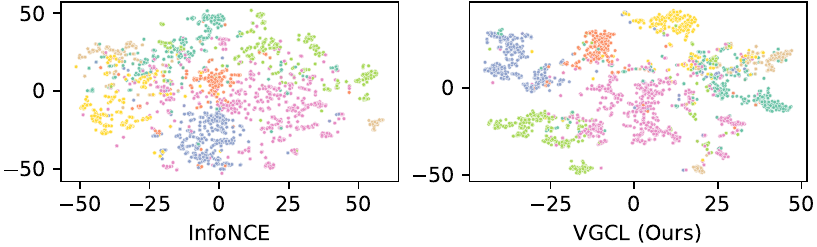}}
\end{figure}

We evaluate the proposed methods on the Planetoid citation datasets for node classification (\cref{apd:exp details}). We use a standard transductive set-up for SSL to evaluate the methods \citep{liu2021}. That is, we train the self-supervised learner on all available data points and afterwards fit a linear classifier on the obtained embeddings. A 10/10/80 split is used and the embeddings are evaluated with the mean test accuracy over different data splits. To compare the usefulness of the generated (unsupervised) uncertainty estimates we use the resulting retention curves of the linear classifier on the test set. Unless specified otherwise, we report the mean and standard error estimates based on 20 runs.

\paragraph{Probabilistic Encoders}

On all datasets, an encoder with Gaussian priors performs on-par or slightly better than its deterministic counterpart (\cref{tab:acc_table}). In addition to that, we find that using variational graph contrastive learning with hyperpriors (VGCL) yields significant performance improvements of up to $1.3$ percent points.  A visualization of the created embeddings with a T-SNE plot shows that these improvements in accuracy stem from a better separation of the embedding space (\cref{fig:tsns}). We also find that setting a hyperprior on the variance of the variational family leads to much better uncertainty estimates. (\cref{fig:ret_curves}, right).

\paragraph{Representation Uncertainty}

\begin{figure}[t]
\floatconts
  {fig:ret_curves}
  {\caption{(Left) Error retention curves on Cora for different uncertainty measures for a deterministic InfoNCE model. We see that our proposed uncertainty measure yields the best sorting. (Right) Performance of the CMDS for different models. We see that the uncertainty calibration improves when Bayesian weight uncertainty is taken into account. To generate the plots, the test data is ordered by increasing uncertainty and the accuracy is calculated as gradually more test samples are incorporated. For a good uncertainty measure the mean accuracy is high for the most certain samples and decreases as more uncertain data points are included.}}
{\includegraphics[width=1\linewidth]{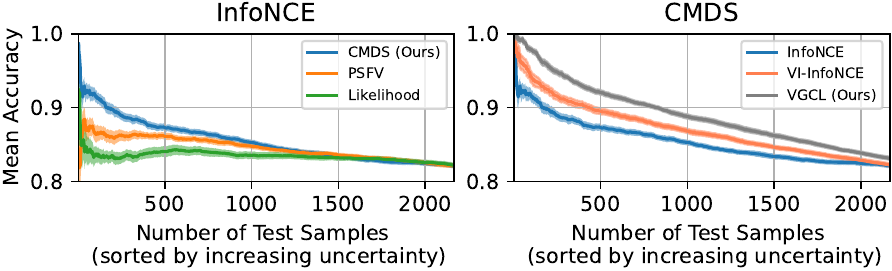}}
\end{figure}

For the trained models, we calculate the average standard deviation of the embedding features \citep[ASTD,][]{BGCL}, the min-max normalized ASTD, the Per-Feature-Sample-Variance  \citep{ardeshir2022uncertainty}, the expected likelihood under positive samples $\mathbb{E}_{x' \sim p(x\mid z')} \left[ \mathrm{P}_{\theta, \rho}(x\mid x') \right]$, the related Watanabe-Akaike Information Criterion (WAIC) $\mathbb{E}_{x' \sim p(x\mid z')} \left[ \mathrm{P}_{\theta, \rho}(x\mid x') \right] - \mathrm{Var}_{x' \sim p(x\mid z')} \left[ \log \mathrm{P}_{\theta,\rho}\left(x\mid x' \right) \right]$, and the CMDS under positive samples as proposed in \cref{sec:cmds}. We consider all scores with and without uncertainty in the weights. We illustrate some representative results in \cref{fig:ret_curves} and show additional ones in \cref{apd:add_res}. We observe that our proposed uncertainty measure yields the best correlation with downstream predictive accuracy of all the considered criteria (\cref{fig:ret_curves}, left) and we also see that our proposed method provides the best uncertainties (\cref{fig:ret_curves}, right).

\section{Conclusion}

We have proposed VGCL, a variational Bayesian approach to graph contrastive learning, and CMDS, a novel measure for uncertainty in the learned representations.
We have shown that VGCL yields better downstream performance than the baseline methods on all investigated graph datasets and that it yields uncertainties that are better calibrated.
Moreover, we have shown that CMDS correlates better with downstream accuracy than other criteria from the literature.
In future work, it would be interested to study whether different priors or variational distributions could additionally improve the performance of VGCL and whether the superior uncertainty estimation of the combination of VGCL with CMDS transfers from the graph setting onto other modalities (e.g., images).

\acks{The work of Elvin Isufi is supported by the TU Delft AI Labs programme. Alexander Immer is supported by a Max Planck ETH Center for Learning Systems doctoral fellowship. Vincent Fortuin was supported by a Branco Weiss Fellowship.}

\bibliography{jmlr-sample}

\newpage

\appendix

\section{Experimental Details}\label{apd:exp details}
\paragraph{Datasets}
We utilize three widely-used citation networks, Cora, Citeseer, Pubmed, for predicting article subject categories \citep{cora_citation,citation_networks}. The Cora dataset consists of 2708 scientific publications that constitute the nodes while CiteSeer is made up of 3327 and Pubmed of 19717. The edges are the citation links between the papers and the features are a sparse bag-of-words for each node.

\paragraph{Models}
As encoder we use a two-layers GCN as proposed by \citet{kipf2016semi}:
\begin{equation}
\begin{aligned}
\mathrm{E}_i(\boldsymbol{X}, \boldsymbol{A}) & =\sigma\left(\hat{\boldsymbol{D}}^{-\frac{1}{2}} \hat{\boldsymbol{A}} \hat{\boldsymbol{D}}^{-\frac{1}{2}} \boldsymbol{X} \boldsymbol{W}_i\right) \\
Z(\boldsymbol{X}, \boldsymbol{A}) & =\mathrm{E}_2\left(\mathrm{E}_1(\boldsymbol{X}, \boldsymbol{A}), \boldsymbol{A}\right) .
\end{aligned}
\end{equation}
where we use the adjacency matrix with self-loops $\hat{\boldsymbol{A}}= \boldsymbol{A} + \boldsymbol{I}$  and normalise with the degree matrix  $\hat{\boldsymbol{D}}=\sum_i \hat{\boldsymbol{A}}_i$. For all models we use a hidden layer size of 128 and a ReLU activation function. In addition to that, we use two MLP layers with a hidden layer size of 128 and an ELU activation as  projection head. We then calculate the cosine similarity of the output as similarity in the contrastive loss. 

\paragraph{Training and Hyperparameters}
We train all models for 150 epochs for the Cora and Citeseer datasets and for 1500 epochs for Pubmed and use Adam for parameter optimization  \citep{kingma2014adam}. Furthermore, we use the temperature-scaled InfoNCE loss as introduced by \citet{SimCLR} and applied to graphs by \citet{GraphCL}. We average the loss over $20$ samples when training the probabilistic encoders. The data is augmented via a mix of feature masking and edge dropping as proposed by \citet{Grace}. We thus create graph augmentations by applying a Bernoulli dropout on the edges and features where we tune the respective probabilities. Following  \citet{liu2021}, during training we create two augmentations of the original graph and compare these via the contrastive loss. Hereby, we tune the Bernoulli dropout probabilities to create the two different augmentations separately resulting in the hyperparameters $p_{f,1},p_{e,1},p_{f,2},p_{e,2}$.  We obtain the accuracies used in the paper with the  hyperparameter settings in \cref{tab:hyp}.

\begin{table}
\centering
\begin{tabular}{ccccccccccc}
\hline Dataset& Model & $p_{f, 1}$ & $p_{f, 2}$ & $p_{r, 1}$ & $p_{r, 2}$ & $\begin{array}{c}\text { lr } \end{array}$  & $\sigma^2$ & $\sigma_{0}$ & $\mu_{p}^2$ & $\sigma_{p}^2$ \\
\hline Cora & VI-InfoNCE  & 0.3 & 0.3 & 0.4 & 0.4 & $5*10^{-3}$  & 0.0025 & - & - & - \\
Citeseer & VI-InfoNCE & 0.3 & 0.4 & 0.4 & 0.4 & $5*10^{-3}$  & 0.0010 & - & - & -\\
Pubmed & VI-InfoNCE  & 0.0 & 0.2 & 0.4 & 0.1 & $10^{-2}$  & 0.0100 & - & - & -\\
Cora & VGCL & 0.3 & 0.3 & 0.4 & 0.4 & $10^{-2}$  & 0.0025 & - & 0 & $10^{-6}$\\
CiteSeer & VGCL & 0.3 & 0.4 & 0.4 & 0.4 & $5*10^{-3}$  & 0.0010 & 0.001 & $10^{-3}$ & $10^{-3}$\\
Pubmed & VGCL & 0.0 & 0.2 & 0.4 & 0.1 & $10^{-2}$  & 0.0100 & - & 0 &  $10$\\
\hline
\end{tabular}
\caption{Hyperparameters}
\label{tab:hyp}
\end{table}

\newpage
\paragraph{Evaluation and Baselines}
As we consider the transductive setting the models are trained on all the available unlabeled data in an unsupervised manner. Afterwards the epoch with the smallest contrastive loss is chosen and the created embeddings are evaluated by fitting a $l_{2}$-regularized logistic regression. Hereby, we follow \citet{liu2021} and use $10\%$ of the data for training,  $10\%$ for validation and $80\%$ for testing. For the Bayesian Models we sample $100$ embeddings for each node and use the average feature values in the evaluation. \\\\
As baselines for the uncertainty quantification task we have considered the measures in the literature that are unsupervised and are directly calculatable for a trained InfoNCE learner (i.e not based on an additional model for density estimation such as a Gaussian-Mixture-Model). In addition to that, we note that the Per-Sample-Feature-Variation \citep[PSFV,][]{ardeshir2022uncertainty} and the Average-Standard-Deviation of the features \citep[ASTD,][]{BGCL} are computationally very similar quantities. Nevertheless, in the PSFV the variation in the embeddings is a result of taking different augmentations, while in the ASTD the uncertainty comes from the distribution over the weights. Therefore, the ASTD score is not applicable to a deterministic encoder. Generally, when we calulate the uncertainty measures in our experiments, we use the contrastive loss to approximate the likelihood. In practice we use the objective that is being minimized but change the order as appropriate for the sorting of the retention curves. The acronyms in the results presented in \cref{tab:acc_table} are Graph Autoencoder, Variational Graph Autoencoder \citep[GAE,VGAE,][]{kipf2016semi} and Deep Graph Infomax \citep[DGI,][]{veličković2018deep}. We have implemented the InfoNCE, VI-InfoNCE and VGCL methods ourselves and have taken the results for the other ones from \citet{Grace} where the same experimental procedure and datasets are used as in our work.

\newpage
\section{Derivation of the ELBO for a probabilistic encoder}\label{apd:der elbo}

In this section, we step through the derivation of the ELBO that includes weight uncertainty in the encoders and in the projection head. When reading the derivation it is important to keep in mind, that the model proposed by \citet{aitchison2023infonce} has little similarity with a variational autoencoder that uses a reconstruction loss. The term variational autoencoder is used in the literal sense, that we learn by optimizing an evidence lower bound. \\

In the model we consider two separate  encoders (usually we share weights and effectively train one neural network), one that encodes $x$ and the other one that encodes $x'$.  Thus, to be fully general, we have to consider the weights over the first network $\phi_{1}$, the weights over the second network $\phi_{2}$ and the weights in the similiarity function $\rho$ separately. For the following derivation we furthermore use that:

\begin{align}
\mathrm{P}\left(x, x^{\prime},z,z^{\prime},\phi_{1},\phi_{2},\rho \right) 
&= \mathrm{P}\left(x, x^{\prime},z,z^{\prime}\mid \phi_{1},\phi_{2}, \rho \right)\mathrm{P}\left( \phi_{1}, \phi_{2}, \rho \right) \\ 
&= \mathrm{P}\left(x|z,\phi_{1} \right)\mathrm{P}\left(x^{\prime}|z^{\prime},\phi_{2} \right)\mathrm{P}\left(z,z^{\prime}|,\phi_{1},\phi_{2},\rho \right)\mathrm{P}\left(\phi_{1},\phi_{2},\rho \right)
\end{align}

which is because the data is modeled as independent given the latents. The likelihoods are approximated with:

\begin{align}
\mathrm{P}\left(x|z,\phi_{1} \right)= \frac{\mathrm{Q}(z \mid x,\phi_{1}) \ \mathrm{P}_{\text {true }}\left(x\right)}{\mathrm{Q}(z|\phi_{1})} 
\end{align}

\begin{align}
\mathrm{P}\left(x'|z',\phi_{2} \right)= \frac{\mathrm{Q}(z' \mid x',\phi_{2}) \ \mathrm{P}_{\text {true }}\left(x'\right)}{\mathrm{Q}(z'|\phi_{2})}
\end{align}

while this definition of the likelihoods might seem unfamiliar at first, it stems from the absence of a reconstruction loss and from the fact that the encoders are not invertible (for further details see \citet{aitchison2023infonce}). In addition, the approximate  posterior for two data points decomposes as:

\begin{equation}
\mathrm{Q}(z,z' \mid x,x',\phi_{1} \ \phi_{2}) = \mathrm{Q}(z \mid x,\phi_{1}) \ \mathrm{Q}(z' \mid x',\phi_{2})
\end{equation}

Then:
\begin{flalign}
 &\log\mathrm{P}\left(x, x^{\prime}\right)  \\ \nonumber
 &=\log \int d\phi_{1} \ d\phi_{2} \ d\rho \ d z \ d z^{\prime} \mathrm{P}\left(x, x^{\prime},z,z^{\prime},\phi_{1}, \phi_{2}, \rho \right) \\\nonumber
&=\log \int  d\phi_{1} \ d\phi_{2} \ d\rho \ d z \ d z^{\prime} \mathrm{P}\left(x, x^{\prime},z,z^{\prime}\mid \phi_{1}, \phi_{2}, \rho \right)\mathrm{P}\left( \phi_{1}, \phi_{2}, \rho \right)\\ \nonumber
&=\log \int d\phi_{1} \ d\phi_{2} \ d\rho  \ d z  \ d z^{\prime} \mathrm{Q}(z,z' \mid x,x',\phi_{1}, \phi_{2}) \ \mathrm{q}(\phi_{1}, \phi_{2}, \rho \mid \theta) \frac{\mathrm{P}\left(x, x^{\prime},z,z^{\prime}\mid \phi_{1},\phi_{2}, \rho \right)\mathrm{P}\left( \phi_{1}, \phi_{2}, \rho \right)}{\mathrm{Q}(z,z' \mid x,x',\phi_{1} \ \phi_{2}) \ \mathrm{q}(\phi_{1},\phi_{2}, \rho \mid \theta)} \nonumber
\end{flalign}

\ \ 

Identifying this as an expectation and plugging in the approximate likelihoods from above we get: 

\begin{align}
& \nonumber \log \mathrm{E}_{\mathrm{Q}\left(z, z^{\prime}\mid x,x^{\prime}, \phi_{1},\phi_{2} \right)\mathrm{q}(\phi_{1},\phi_{2}, \rho \mid \theta)}\left[\frac{\mathrm{Q}(z' \mid x',\phi_{2}) \ \mathrm{P}_{\text {true }}\left(x'\right)\mathrm{Q}(z \mid x,\phi_{1}) \ \mathrm{P}_{\text {true }}\left(x\right)\mathrm{P}\left(z,z^{\prime}\mid \phi_{1},\phi_{2}, \rho \right)\mathrm{P}\left( \phi_{1},\phi_{2}, \rho \right)}{\mathrm{Q}(z,z' \mid x,x',\phi_{1} \ \phi_{2}) \mathrm{Q}\left(z^{\prime}\mid \phi_{2} \right)\mathrm{Q}\left(z \mid \phi_{1} \right ) \ \mathrm{q}\left(\phi_{1},\phi_{2}, \rho \mid \theta\right)}\right]  \\ \nonumber
&=\log \mathrm{E}_{\mathrm{Q}\left(z, z^{\prime}\mid x,x^{\prime}, \phi_{1},\phi_{2} \right)\mathrm{q}(\phi_{1},\phi_{2}, \rho \mid \theta)}\left[\frac{\mathrm{P}\left(z,z^{\prime}\mid \phi_{1},\phi_{2}, \rho \right)\mathrm{P}\left( \phi_{1},\phi_{2}, \rho \right)}{\mathrm{Q}\left(z^{\prime}\mid \phi_{2} \right)\mathrm{Q}\left(z \mid \phi_{1} \right) \ \mathrm{q}\left(\phi_{1},\phi_{2}, \rho \mid \theta \right)}\right] + \text{const} \\ \nonumber 
&\geq  \mathrm{E}_{\mathrm{Q}\left(z, z^{\prime}\mid x,x^{\prime},\phi_{1},\phi_{2} \right)\mathrm{q}(\phi_{1},\phi_{2},\rho  \mid \theta)}\left[\log\frac{\mathrm{P}\left(z,z^{\prime}\mid \phi_{1},\phi_{2}, \rho \right)\mathrm{P}\left( \phi_{1},\phi_{2}, \rho \right)}{\mathrm{Q}\left(z^{\prime}\mid \phi_{2} \right)\mathrm{Q}\left(z \mid \phi_{1} \right) \ \mathrm{q}\left(\phi_{1},\phi_{2}, \rho \mid \theta\right)} \right] + \text{const}\\  \nonumber
\end{align}

Now, if we model the weights $\phi_{1},\phi_{2},\rho$ as independent from each other the expression we want to optimize turns out to be: 

\begin{align}
 \log\mathrm{P}\left(x, x^{\prime}\right) \geq \
  &\mathrm{E}_{\mathrm{Q}\left(z, z^{\prime}\mid x,x^{\prime},\phi_{1},\phi_{2} \right)\mathrm{q}(\phi_{1},\phi_{2},\rho  \mid \theta)}\left[\log\frac{\mathrm{P}\left(z,z^{\prime}\mid \phi_{1},\phi_{2}, \rho \right)}{\mathrm{Q}\left(z^{\prime}\mid \phi_{2}\right)\mathrm{Q}\left(z \mid \phi_{1}\right)} \right] \\
 & - KL(\mathrm{q}(\phi_{1}\mid \theta)||P(\phi_{1}))  \\
 & -KL(\mathrm{q}(\phi_{2} \mid \theta)||P(\phi_{2})) \\
 & - KL(\mathrm{q}(\rho \mid \theta)||P(\rho)) \\
 & + \text{const}
\end{align}

\newpage
\section{The InfoNCE as a prior over the embeddings}\label{apd:InfoNCE_ELBO}

In this section we lay out how for a deterministic encoder the prior given in the methods section results in the infinite-sample InfoNCE objective. For a deterministic encoder we can derive the ELBO analogous to the previous section, but without introducing the variational family $\mathrm{q}(\phi_{1},\phi_{2},\rho  \mid \theta)$ and the distribution $\mathrm{P}\left( \phi_{1},\phi_{2}, \rho \right)$. For ease of exposition, we furthermore assume that we use the same neural network to encode both data points $x$ and $x'$ with weights $\phi$. The resulting ELBO then turns out to be:

\begin{align}
 \log\mathrm{P}\left(x, x^{\prime}\right) \geq \
  &\mathrm{E}_{\mathrm{Q}_{\phi}\left(z, z^{\prime}\mid x,x^{\prime}\right)}\left[\log\frac{\mathrm{P}_{\phi,\rho}\left(z,z^{\prime}\right)}{\mathrm{Q}_{\phi}\left(z^{\prime}\right)\mathrm{Q}_{\phi}\left(z \right)} \right] + \text{const}
\end{align}

Now, averaging over the data points this expression turns into:

\begin{align}
  \mathrm{E}_{\mathrm{Q}_{\phi}\left(z, z^{\prime}\right)}\left[\log\frac{\mathrm{P}_{\phi,\rho}\left(z,z^{\prime}\right)}{\mathrm{Q}_{\phi}\left(z^{\prime}\right)\mathrm{Q}_{\phi}\left(z \right)} \right] + \text{const}
\end{align}

and then we plug in the prior:

\begin{equation}
\begin{aligned}
\mathrm{P}_{\phi,\rho}^{\operatorname{InfoNCE}}(z) & =\mathrm{Q}_\phi(z) \\
\mathrm{P}_{\phi, \rho}^{\operatorname{InfoNCE}}\left(z^{\prime} \mid z\right) & =\frac{1}{Z_{\phi, \rho}(z)} \mathrm{Q}_\phi\left(z^{\prime}\right) f_\rho\left(z, z^{\prime}\right)
\end{aligned}
\end{equation}

with the normalising constant: 

\begin{equation}
Z_{\rho, \phi}(z) = \int \mathrm{Q}_\phi\left(z^{\prime}\right) f_\rho\left(z, z^{\prime} \right)d z^{\prime}
\end{equation}

 Note that this prior is parameterised in a way that it shares the parameters $\phi$ with the encoder. This is a trick frequently used in the literature  \citep{zhao2018information,ustyuzhaninov20a}. We can then obtain the ELBO for the InfoNCE:

\begin{equation}
\mathcal{L}_{\text {InfoNCE }}(\theta, \phi)=\mathrm{E}_{\mathrm{Q}_\phi\left(z, z^{\prime}\right)}\left[\log \frac{\mathrm{Q}_\phi(z) \frac{1}{Z_{\theta, \phi}(z)} \mathrm{Q}_\phi\left(z^{\prime}\right) f_\theta\left(z, z^{\prime}\right)}{\mathrm{Q}_\phi(z) \mathrm{Q}_\phi\left(z^{\prime}\right)}\right] + \text{const}
\end{equation}

by cancelling we get: 

\begin{equation}
\mathcal{L}_{\text {InfoNCE }}(\theta, \phi)=\mathrm{E}_{\mathrm{Q}_\phi\left(z, z^{\prime}\right)}\left[\log \frac{f_\theta\left(z, z^{\prime}\right)}{Z_{\theta, \phi}(z)}\right] + \text{const}
\end{equation}

at this point we already note the similarity with \cref{eqn:infonce}, which is the finite-sample estimator. This can also be seen formally by plugging in $Z_{\theta, \phi}(z)$ and rewriting the resulting expression as:

\begin{equation}
 \mathcal{L}_{\text {InfoNCE }}(\theta, \phi)=\mathrm{E}_{\mathrm{Q}_\phi\left(z, z^{\prime}\right)}\left[\log f_\theta\left(z, z^{\prime}\right)\right] 
-\mathrm{E}_{\mathrm{Q}_\phi(z)}\left[\log \mathrm{E}_{\mathrm{Q}_\phi\left(z^{\prime}\right)}\left[f_\theta\left(z, z^{\prime}\right)\right]\right]+\text { const. }
\end{equation}

which is up to a constant equivalent to the infinite-sample InfoNCE objective \citep{wang2020understanding,li2021self}.

\section{Motivation for the CMDS}\label{apd:Dis Score}

In this section we will lay out and further interpret the Contrastive Model Disagreement Score (CMDS) proposed in the main paper. Recall that it is defined as: 

 \begin{align}
D_{\text{CMDS}}\left(x\right)=\frac{1}{M \sum_{j=1}^{M} l_{j}(x)^2} \, , \qquad \text{with} \quad l_{j}(x)=\frac{\mathrm{P}\left(x \mid x'_{j}, \phi_{j}, \rho_{j} \right)}{\sum_{j=1}^{M} \mathrm{P}\left(x \mid x'_{j}, \phi_{j}, \rho_{j} \right)}
\end{align}

Hereby, we first obtain $M$ different likelihoods for the data point $x$ that we are interested in by sampling different positive augmentations and weights (if a BNN is used). Then we normalise the likelihoods to obtain the set $\{l_{j}(x)\}_{j=1}^{M}$. This normalisation is important, because otherwise the variaton between these likelihoods would depend on their absolute size - which would make it unsuitable for our purposes. We then use the normalised likelihoods to calculate the score $D_{\text{CMDS}}(x) \in [\frac{1}{M},1]$ which obtains its maximum when the $\{l_{j}(x)\}_{j=1}^{M}$ are uniformly distributed. In contrast, when the likelihoods are very different with a few comparably large values then $D_{\text{CMDS}}(x)$ will be small. In the extreme case, when we have $\{l_{j}(x)\}_{j=1}^{M} = \{0,...,0,1,0,...,0\}$, it is equal to $\frac{1}{M}$. To see that this is indeed the minimum we note that $0 \leq l_{j} \leq 1$ and thus $M \sum_{j=1}^{M} l_{j}(x)^2 \leq M \sum_{j=1}^{M} l_{j}(x) = M$. To show that $1$ is the maximum value we can use Cauchy Schwartz $(\sum_{j=1}^{M} l_{j}(x) \ y_j)^2 \leq(\sum_{j=1}^{M} l_{j}(x)^2) \ (\sum y_{j}^{2} )$ and let $y_i =1$. This gives 
$(\sum_{j=1}^{M} l_{j}(x))^2 \leq \sum_{j=1}^{M} l_{j}(x)^2 M$ and because $\sum_{j=1}^{M} l_{j}(x) = 1$ we have $\frac{1}{M} \leq \sum_{j=1}^{M} l_{j}(x)^2$. 
\\\\
The $D_{\text{CMDS}}(x)$ has an intrinsic connection to information theory. To illustrate this, suppose that we have observed some data and inferred the posterior distribution of the model parameters $\mathrm{P}(\phi, \rho \mid \mathrm{D} )$. This quantity would change if we observed some new data $x$ and $x'$ and would update the parameters to $\mathrm{P}(\phi, \rho \mid \mathrm{D}^{*})$. This change would naturally be bigger the more the new observed data points differ from the data $\mathrm{D}$ that the model had already seen. In fact, the formulation that we use for the disagreement score can be viewed as a measure that quantifies this change in distribution or ,equivalently, the informativeness of observing the pair of data points $x$ and $x'$ \citep{BVAE}. We can thus view the $D_{\text{CMDS}}(x)$ as a score of informativeness of observing a data point $x$ in a contrast given by samples from $p(x'\mid z')$. If $x$ is very similar to these samples then the additional information is limited and the model has a coherent explanation for observing $x$ given that contrast. In that scenario, there is agreement between the likelihoods and  $D_{\text{CMDS}}(x)$ is large.  \\\\
This becomes even clearer when we consider a different perspective on augmentations and note that we could theoretically shift the Bernoulli distributions from the graph to the weights of the neural network. For example,  consider the scenario in which we encode a positive sample data point $x'$, created by a Bernoulli dropout on the features of $x$. This  can be interpreted as directly encoding the original data point $x$, but with an altered (dropout) distribution in the first layer weights of the encoding neural network. When looking at it from that perspective, $D_{\text{CMDS}}(x)$ can be interpreted similarly to   the disagreement score over Bayesian weight distributions for variational autoencoders as proposed by \citet{BVAE}.

\section{Properties of the Disagreement Score}\label{apd:Uncertainty Measure}

In this section we show that $P(x \mid x')$ is independent of the true data distributions $P_{true}(x')$ and $P_{true}(x)$. This is a useful property of the CMDS as it makes it robust to distribution shifts and likely to work across datasets. To show this we follow a similar procedure as in \cref{apd:der elbo} : 

\begin{align}
\log \mathrm{P}_{\phi,\rho}\left(x\mid x' \right) &=\log \int d z  d z^{\prime} \mathrm{P}_{\phi,\rho}\left(x \mid x',  z,z^{\prime} \right)  \mathrm{P}_{\phi,\rho}\left(z,z^{\prime} \mid x'  \right)  \\ &= 
\log \int d z  d z^{\prime} \mathrm{P}_{\phi,\rho}\left(x \mid z \right)   \mathrm{P}_{\phi,\rho}\left(z \mid z^{\prime} \right) \mathrm{P}_{\phi,\rho}\left(z^{\prime} \mid x'  \right)
\end{align}

where we used that $x$ is independent of $x', z'$ when conditioned on its generating latent $z$ and similiar that $z$ is independent of $x'$ when $z'$ is given. Plugging in the approximate likelihood for $x$, the InfoNCE prior and the approximate posterior for $x'$ we get: 

\begin{align}
\log \mathrm{P}_{\phi,\rho}\left(x\mid x' \right) &= \log \int d z  d z^{\prime} \frac{\mathrm{P}_{\phi,\rho}\left(x \mid z \right)   \mathrm{Q}_\phi\left(z\right) f_\rho\left(z', z \right)\mathrm{Q}_{\phi}\left(z' \mid x_{obs}  \right)}{Z_{\phi, \rho}(z)} \\ 
 &= \log \int d z  d z^{\prime} \frac{\mathrm{Q_{,\phi}}(z \mid x) \ \mathrm{P}_{\text {true }}\left(x\right)\mathrm{Q}_\phi\left(z\right) f_\rho\left(z', z \right)\mathrm{Q}_{\phi}\left(z' \mid x_{obs}  \right)}{\mathrm{Q}_{\phi}\left(z \right)Z_{\phi, \rho}(z)} \\
  &= \log \int d z  d z^{\prime} \frac{\mathrm{Q_{,\phi}}(z \mid x)  f_\rho\left(z', z \right)\mathrm{Q}_{\phi}\left(z' \mid x'  \right)}{Z_{\phi, \rho}(z)} + \log \mathrm{P}_{\text {true }}\left(x\right)
\end{align}

Where we can notice that the term on the left is really just the InfoNCE that we obtain for $x$ and $x'$ but by conditioning we have gotten rid of the dependence on the true underlying distribution $\mathrm{P}_{\text {true }}\left(x' \right)$. For ease of exposition we will call the first part of the sum which is the contrastive loss $K$. We the have:

\begin{align}
\mathrm{P}_{\phi,\rho}\left(x\mid x' \right) = K_{\phi,\rho} * P_{true}(x)
\end{align}

Plugging this into the discriminator we get for each $l_{i}$:

\begin{align}
l_{i}=\frac{\mathrm{P}\left(\mathbf{x} \mid x'_{i}, \phi_{i}, \rho_{i} \right)}{\sum_{i=1}^{N} \mathrm{P}\left(\mathbf{x} \mid x'_{i}, \phi_{i}, \rho_{i} \right)} = \frac{K_{\phi_{i},\rho_{i}}\mathrm{P}_{\text {true }}\left(x \right)}{\sum_{i=1}^{N} K_{\phi_{i},\rho_{i}}\mathrm{P}_{\text {true }}\left(x \right)} =\frac{K_{\phi_{i},\rho_{i}}}{\sum_{i=1}^{N} K_{\phi_{i},\rho_{i}}}
\end{align}

where the latter term is independent of the true data distributions $\mathrm{P}_{\text {true }}\left(x' \right)$,$\mathrm{P}_{\text {true }}\left(x \right)$. 

\newpage
\section{Additional Results}\label{apd:add_res}

In this section we present and discuss some additional results related to the methods presented in the paper. In general, we find that the CMDS provides a better calibrated measure of uncertainty than the baselines across models and datasets (\cref{fig:ret_curves_pubmed_models,fig:ret_curves_cora_models,fig:ret_curves_citeseer_models,fig:ret_curves}). While we also find that incorporating weight uncertainty is generally beneficial, it impacts some measures of uncertainty more than others. Notably, when using variational inference the performance of the likelihood related measures (Likelihood, WAIC, CMDS) consistently outperforms the ones that focus on the embedding space (ASTD, ASTD\_Norm, PSVF) as can be seen in \cref{fig:ret_curves_pubmed_models,fig:ret_curves_cora_models,fig:ret_curves_citeseer_models} (right). One reason for this difference in improvement could be, that the first are able to incorporate the epistemic uncertainty in the projection head, while the second can only make use of the epistemic uncertainty in the encoders. That being said, the PSVF outperforms the Likelihood and the WAIC in a deterministic setting (\cref{fig:ret_curves_pubmed_models,fig:ret_curves}, left). \\

Furthermore, it seems that the measures do not equally well identify the most- and least-uncertain data points. While the measures that are based on the variation of embeddings seem to be better at identifying the most confident examples, the Likelihood and WAIC seem to be useful for the least confident ones (\cref{fig:ret_curves_citeseer_models,fig:ret_curves_cora_models}, left). Generally, using the VGCL model yields the best calibrated uncertainties. This is fundamentally caused by an increase in the variance over the weights. To illustrate this, we have depicted the learned first-layer weight distributions of a trained VGCL in \cref{fig:paramsnetwork} with increasing regularization. We have also added a heatmap over the parameters of an VI-InfoNCE model (\cref{fig:vi_inf_heatmap}).

\begin{figure}[!h]
 \floatconts
  {fig:ret_curves_pubmed_models}
  {\caption{ (Left) Retention Curves on Pubmed for a InfoNCE model. (Right) Retention Curves on Pubmed for VI-InfoNCE.  }}
  {\includegraphics[width=1\linewidth]{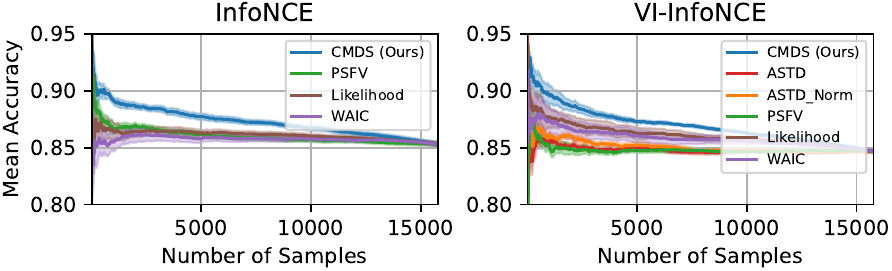}}
\end{figure}

\begin{figure}[!h]
\floatconts
  {fig:ret_curves_cora_models}
  {\caption{ (Left) Retention Curves on Cora for VI-InfoNCE. (Right) Retention Curves on Cora for VGCL.  }}
  {\includegraphics[width=1\linewidth]{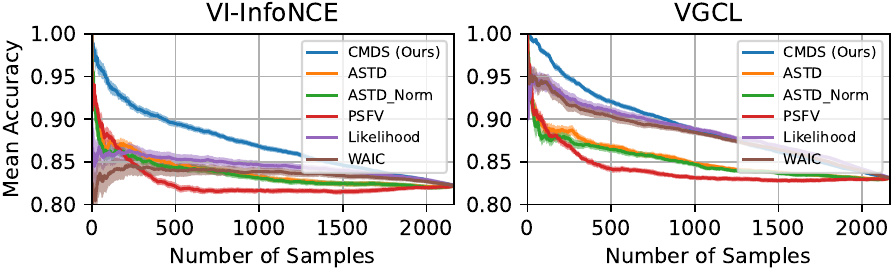}}
\end{figure}

\begin{figure}[!h]
\floatconts
  {fig:ret_curves_citeseer_models}
  {\caption{ (Left) Retention Curves on CiteSeer for VI-InfoNCE. (Right) Retention Curves on CiteSeer for VGCL.  }}
  {\includegraphics[width=1\linewidth]{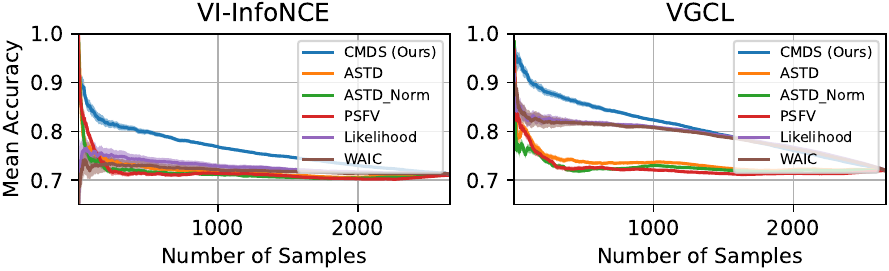}}
\end{figure}

\begin{figure}[!h]
\floatconts
  {fig:vi_inf_heatmap}
  {\caption{Heatmap over the parameters of a VI-InfoNCE model on Cora}}
  {\includegraphics[width=0.5\linewidth]{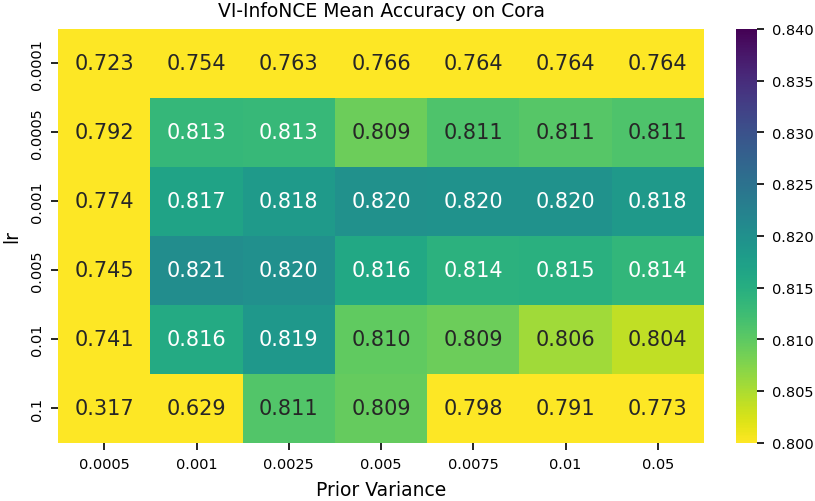}}
\end{figure}

\begin{figure}[!h]
\floatconts
  {fig:paramsnetwork}
  {\caption{Weight parameters in the first layer of VGCL under different regularizations of the variance of the variational family. The stronger the influence of the hyperprior, the larger the standard deviation of the weight distributions.  }}
  {\includegraphics[width=0.5\linewidth]{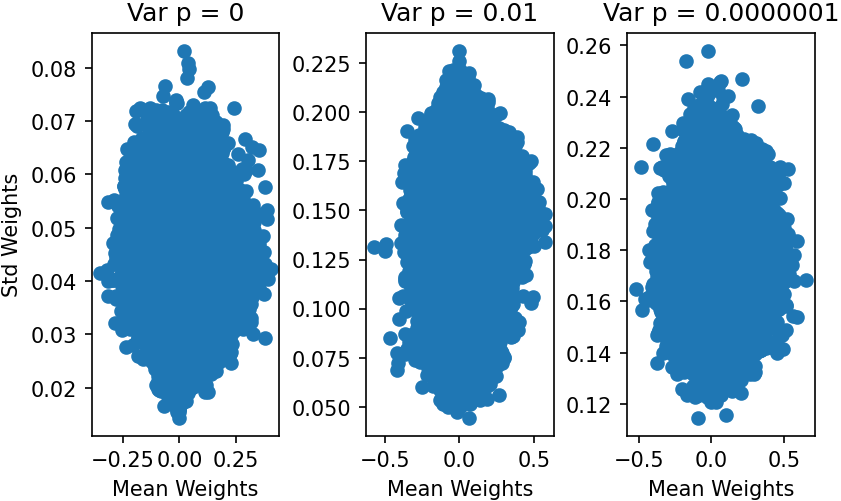}}
\end{figure}

\newpage
\section{Related Work}
\label{sec:related_work}

\paragraph{Uncertainty in Contrastive Learning}

Existing works on uncertainty in contrastive learning generally focus on measuring embedding uncertainty. They do this by using the trained encoder to map the data point in question into the embedding space and then considering different measures of uncertainty. \citet{ardeshir2022uncertainty} measure the variance of the embeddings under different augmentations and \citet{BGCL} additionally consider the uncertainty in the weights of the encoder. The latter then use the average standard deviation (ASTD) of the features of the embeddings. Sometimes, surrogate distributions such as the one obtained from a Gaussian Mixture Model on the embeddings are used to obtain a probability density that can be reasoned about \citep{ardeshir2022uncertainty,wu2020simple}. Other papers try to approximate a notion of hardness of examples by changing a scalar in the similarity function (the temperature) based on intuition and empirical observations \citep{zhang2021temperature}. Note that, \citet{BGCL} and \citet{sharma2023incorporating} also perform Bayesian inference in a contrastive learning setting. The former use this to improve the network performance by learning the parameters of the augmentations and do not learn the weights in the projection head. They use Beta-Bernoulli priors and propose to measure the uncertainty in the embedding space using the ASTD. In contrast to that, we propose to measure uncertainty with regard to the likelihood of the probabilistic model and propose the CMDS. We use the ASTD as a baseline in our experiments and outperform it by a large margin. Furthermore, our improvements in performance in accuracy are not due to improved augmentations, but due to a better specification of the variational family. The latter do not consider learning on graphs and they also only perform Bayesian inference over the last layer of the network, as opposed to the full-network inference in our work.

\paragraph{Bayesian neural networks}

Bayesian neural networks promise to marry the expressivity of neural networks with the principled statistical properties of Bayesian inference \citep{mackay1992practical, neal1993bayesian}.
However, approximate inference in these complex models has remained challenging \citep{jospin2022hands}.
Approximate inference techniques lie on a spectrum of quality and computational cost, from cheap local approximations like Laplace inference \citep{laplace1774memoires, mackay1992practical, khan2019approximate, daxberger2021laplace}, stochastic weight averaging \citep{izmailov2018averaging, maddox2019simple}, and dropout \citep{gal2016dropout, kingma2015variational}, via variational approximations with different levels of complexity \citep[e.g.,][]{graves2011vi, blundell2015bbb, louizos2016structured, khan2018vogn, osawa2019practical}, across ensemble-based methods \citep{lakshminarayanan2017simple, wang2019function, wilson2020bayesian, ciosek2020conservative, he2020bayesian, d2021stein, d2021repulsive}, up to the very expensive but asymptotically correct Markov Chain Monte Carlo (MCMC) approaches \citep[e.g.,][]{neal1993bayesian, neal2011mcmc, welling2011bayesian, garriga2021exact, izmailov2021hmc}.
Apart from the challenges relating to approximate inference, recent work has also studied the question of prior choice for BNNs \citep[e.g.,][and references therein]{fortuin2021bnnpriors, fortuin2022bayesian, nabarro2022data, sharma2023incorporating, fortuin2022priors} and how to perform model selection in this framework \citep[e.g.,][]{immer2021scalable, immer2022invariance, rothfuss2021pacoh, rothfuss2022pac, van2022learning, schwobel2022last}.
In our work, we apply these methods to the graph contrastive learning setting, which to the best of our knowledge has not been studied with full-network variational inference before (see also discussion above).

\paragraph{Graph Contrastive Learning}

In recent years, methods that facilitate learning on graphs have made rapid progress and have been applied to a large variety of problem settings \citep[e.g.,][]{kipf2016semi,xu2018powerful,GTgraph_elv,edge_nets_elv}. Graph contrastive learning (GCL) applies these advances to unlabeled data by using positive and negative examples. Hereby, the contrastive loss is typically based on the InfoNCE (InfoMax) principle, which aims to maximize the mutual information between the input data and its corresponding latent representation \citep{GraphCL,Oord2018,Grace}. Most existing research focuses on improving the performance of GCL by designing methods that produce better augmentations \citep[e.g.,][]{aug3_bag, aug_back1, aug_back_2}, improve negative sampling \citep[e.g.,][]{neg_samp_back1, neg_samples_back2, Grace} or change the loss function \citep[e.g.,][]{back_loss, aug_back1}. In contrast to that, the improvements in accuracy in our work stem from introducing a distribution over the weights.
\end{document}